\documentclass[10pt,twocolumn,english,twocolumn]{article}
\usepackage[T1]{fontenc}
\usepackage[utf8]{inputenc}
\usepackage[a4paper]{geometry}
\usepackage{babel}
\usepackage{array}
\usepackage{booktabs}
\usepackage{graphicx}
\usepackage[unicode=true]
 {hyperref}

\makeatletter

\providecommand{\tabularnewline}{\\}

\newcommand{\lyxaddress}[1]{
\par {\raggedright #1
\vspace{1.4em}
\noindent\par}
}

\@ifundefined{date}{}{\date{}}
\usepackage[font=small,labelfont=bf]{caption}
\makeatletter
\g@addto@macro\@floatboxreset\centering
\makeatother

\makeatother

\begin{document}
\twocolumn[   \begin{@twocolumnfalse}

\title{SMILES Enumeration as Data Augmentation for Neural Network Modeling
of Molecules}

\author{Esben Jannik Bjerrum\textsuperscript{{*}}}

\maketitle

\lyxaddress{Wildcard Pharmaceutical Consulting, Frødings Allé 41, 2860 Søborg,
Denmark\\
\textsuperscript{{*}}) esben@wildcardconsulting.dk}
\begin{abstract}
Simplified Molecular Input Line Entry System (SMILES) is a single
line text representation of a unique molecule. One molecule can however
have multiple SMILES strings, which is a reason that canonical SMILES
have been defined, which ensures a one to one correspondence between
SMILES string and molecule. Here the fact that multiple SMILES represent
the same molecule is explored as a technique for data augmentation
of a molecular QSAR dataset modeled by a long short term memory (LSTM)
cell based neural network. The augmented dataset was 130 times bigger
than the original. The network trained with the augmented dataset
shows better performance on a test set when compared to a model built
with only one canonical SMILES string per molecule. The correlation
coefficient R2 on the test set was improved from 0.56 to 0.66 when
using SMILES enumeration, and the root mean square error (RMS) likewise
fell from 0.62 to 0.55. The technique also works in the prediction
phase. By taking the average per molecule of the predictions for the
enumerated SMILES a further improvement to a correlation coefficient
of 0.68 and a RMS of 0.52 was found.

\bigskip{}

\hrule

\medskip{}

\end{abstract}

 \end{@twocolumnfalse} ]

\section{Introduction}

Neural networks and deep learning has shown interesting application
successes, such as image classification\cite{Simard2003}, and speech
recognition\cite{Hinton2012}. One of the issues that limits their
general applicability in the QSAR domain may be the limited sizes
of the labeled datasets available, although successes do appear.\cite{Mayr2016}
Limited datasets necessitates harsh regularization or shallow and
narrow architectures. Within image analysis and classification, data
augmentation techniques has been used with excellent results.\cite{Almasi2016,Chatfield2014,Cui2015,Schmidhuber2015}
As an example, a dataset of labeled images can be enlarged by operations
such as mirroring, rotation, morphing and zooming. The afterwards
trained network gets more robust towards such variations and the neural
network can recognize the same object in different versions.

Neural networks has also been used on molecular data, where the input
may be calculated descriptors,\cite{Mayr2016} neural network interpretation
of the molecular graph\cite{Kearnes2016a} or also SMILES representations.\cite{Gomez-Bombarelli2016}
Simplified Molecular Input Line Entry System (SMILES) is a single
line text based molecular notation format.\cite{Weininger1970} A
single molecule has multiple possible SMILES strings, which has led
to the definition of a canonical SMILES,\cite{OBoyle2012} which ensures
that a molecule corresponds to a single SMILES string. The possibilities
for variation in the SMILES strings of simple molecules are limited.
Propane has two possibilities CCC and C(C)C. But as the molecule gets
larger in size and more complex in branching, the number of possible
SMILES strings grows rapidly. Toluene with seven atoms, has seven
possible SMILES strings (Figure 1).
\begin{figure}
\begin{centering}
\includegraphics{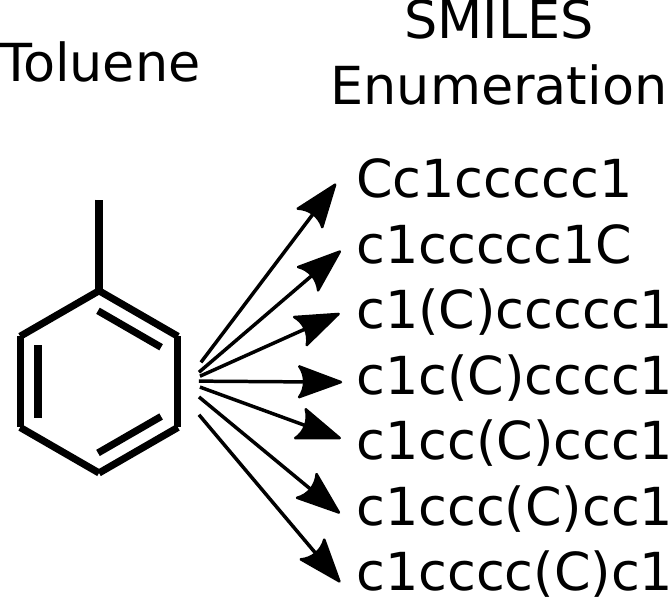}
\par\end{centering}

\caption{SMILES enumeration enables data augmentation. The molecule toluene
corresponds to seven different SMILES, the top one is the canonical
smile. One data point with toluene in the dataset would thus leads
to seven samples in the augmented dataset.}
\end{figure}

Here data augmentation of molecular structures with SMILES enumeration
for QSAR studies will be investigated using long short term memory
(LSTM) cell neural networks inspired by networks used for Twitter
tweets sentiment analysis.\cite{Tang2015}

\section{Methods}

\subsection{SMILES enumeration}

SMILES enumeration was done with a Python script utilizing the cheminformatics
library RDKit.\cite{Landrum2016} The atom ordering of the molecule
is scrambled randomly by converting to molfile format\cite{ctfile2012}
and changing the atom order, before converting back to the RDKit mol
format. A SMILES is then generated using RDKit with the default option
of producing canonical SMILES set to false, where different atom orderings
lead to different SMILES. The SMILES strings is then compared and
possible added to a growing set of unique SMILES strings. The process
is repeated a predefined number of times. The python functions are
available on github: \href{https://github.com/Ebjerrum/SMILES-enumeration}{https://github.com/Ebjerrum/SMILES-enumeration}

\subsection{Molecular dataset }

The dataset was obtained from Sutherland et al 2003.\cite{Sutherland2003a}
It consists of 756 dihydrofolate inhibitors with P. carinii DHFR inhibition
data. The dataset was split in test and a training set in a 1:9 ratio.
It was expanded with SMILES enumeration and the SMILES strings were
padded with spaces to fixed length of 74, which is one characters
longer than the longest SMILES in the dataset. It was subsequently
vectorized by one-hot encoding the characters into a bit matrix with
one bit set for the corresponding character in each row using a generated
char to int dictionary. Molecules where the associated affinity was
not a number were removed. The associated IC50 data was converted
to log IC50 and normalized to unit variance and mean zero with utilities
from Scikit-learn.\cite{scikit-learn}

\subsubsection{LSTM neural network}

Two different neural networks were built and trained using Keras version
1.1.2\cite{chollet2015} with Theano v. 0.8.0\cite{2016arXiv160502688full}
as back end. One or more LSTM layers were used in batch mode, and
the final state fed to a feed-forward neural network with a single
linear output neuron. The network layout was optimized using Bayesian
optimization with Gaussian processes as implemented in the Python
package GpyOpt\cite{gpyopt2016} version 1.0.3, varying the hyper
parameters listed in Table 1. 10 initial trainings was done before
using the GP\_MCMC and the EI\_MCMC acquisition function to sample
new hyper parameter sets.\cite{Wang2013} One network was optimized
and trained only using a dataset with canonical SMILES, whereas the
other were optimized and trained with the dataset that expanded with
SMILES enumeration. In the rest of the publication they will be referred
to as the canonical model and enumerated model, respectively.

\begin{table*}
\caption{Hyper parameter Search Space}

\centering{}%
\begin{tabular}{ccc}
\hline 
\multicolumn{1}{c}{Parameter} & Search Space & Type\tabularnewline
\hline 
Number of LSTM layers & {[}1,2{]} & Discrete\tabularnewline
Number of units in LSTM layers & {[}32, 64, 128, 256{]} & Discrete\tabularnewline
Dropout for input gates (dropout\_W) & 0 – 0.2 & Continuous\tabularnewline
Dropout for recurrent connection (dropout\_U) & 0 – 0.5 & Continuous\tabularnewline
Number of dense hidden layers & {[}0,1{]} & Discrete\tabularnewline
Hidden layer size & {[}4, 8, 16, 32, 64, 128{]} & Discrete\tabularnewline
Weight regularization on dense layer, L1 & 0 – 0.2 & Continuous\tabularnewline
Weight regularization on dense layer, L2 & 0 – 0.2 & Continuous\tabularnewline
Learning rate & 0.05-0.0001 & Continuous\tabularnewline
\hline 
\end{tabular}
\end{table*}

All computations and training were done on a Linux workstation (Ubuntu
Mate 16.04) with 4 GB of ram, i5-2405S CPU @ 2.50GHz and an Nvidia
Geforce GTX1060 graphics card with 6 GB of ram.

\section*{Results}

Filtering, splitting and SMILES enumeration resulted in a canonical
SMILES dataset with 602 train molecules and 71 test molecules, whereas
the enumerated dataset had 79143 and 9412 rows for train and test,
respectively. This corresponds to an augmentation factor of approximately
130. Each molecule had on average 130 alternative SMILES representations.
Optimization of the architecture yielded two different best configurations
of hyper parameters, depending on the dataset used. The best hyper
parameters found for each dataset are shown in Table 2.

\begin{table*}

\caption{Best Hyperparameters found}

\begin{centering}
\begin{tabular}{ccc}
\toprule 
Parameter & Canonical Model & Enumerated Model\tabularnewline
\midrule
Number of LSTM layers & 1 & 1\tabularnewline
Number of units in LSTM layers & 128 & 64\tabularnewline
Dropout for input gates (dropout\_W) & 0.0 & 0.19\tabularnewline
Dropout for recurrent connections (dropout\_U) & 0.0 & 0.0\tabularnewline
Number of dense hidden layers & 0 & 0\tabularnewline
Hidden layer size & N/A & N/A\tabularnewline
Weight regularization on dense layer, L1 & 0.2 & 0.005\tabularnewline
Weight regularization on dense layer, L2 & 0.2 & 0.01\tabularnewline
Learning rate & 0.0001 & 0.005\tabularnewline
\bottomrule
\end{tabular}
\par\end{centering}

\end{table*}

The train history is shown in Figure 2.
\begin{figure*}
\begin{centering}
\includegraphics[width=1\textwidth]{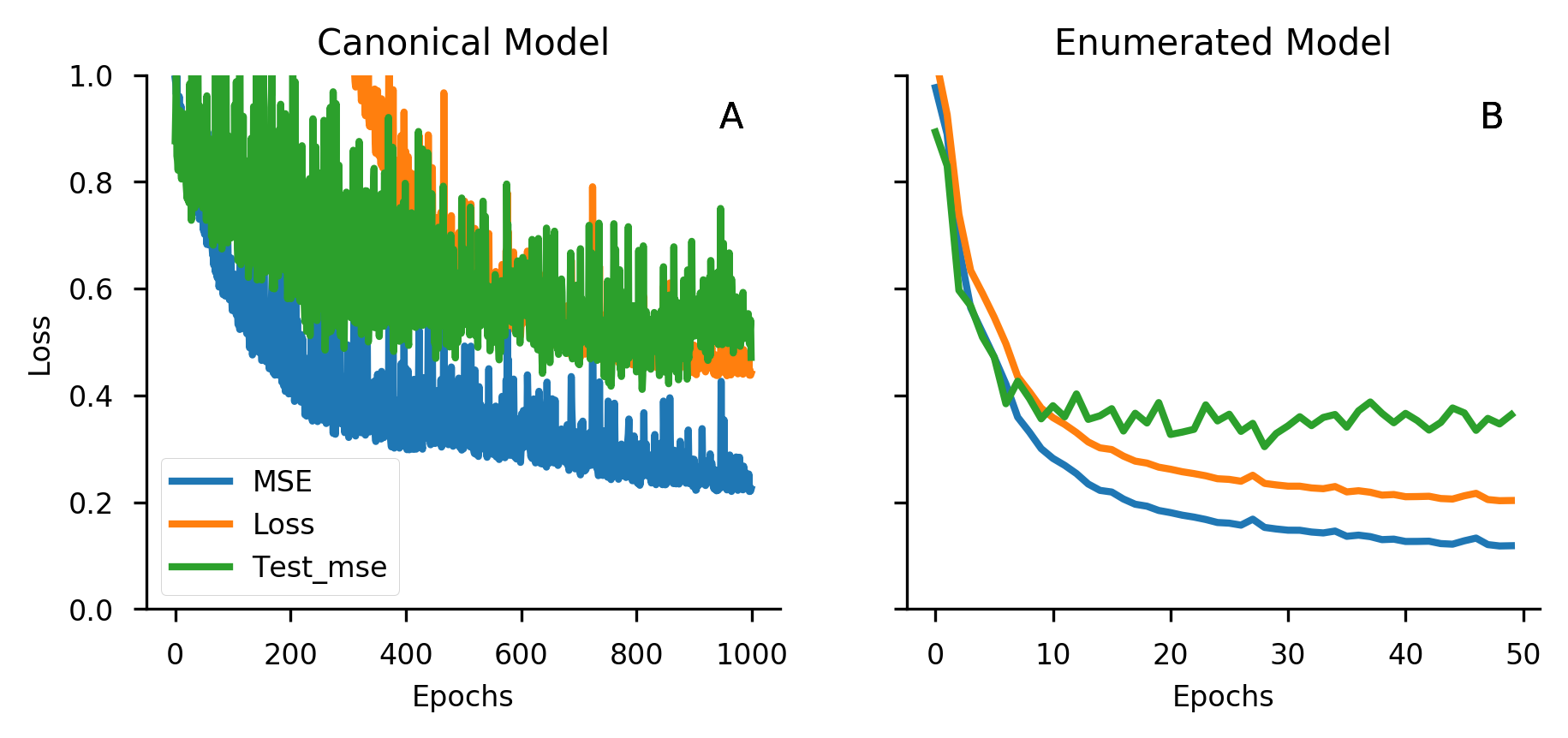}
\par\end{centering}

\caption{Training history for the two datasets and neural networks. A: Neural
network trained on canonical SMILES shows a noisy curve where the
best model has a test loss of 0.41. B: Neural network trained on enumerated
SMILES obtains the best model with a test loss of 0.30. Blue lines
are the mean square error without regularization penalty, green is
loss including regularization penalty and the red line is mean square
error on the test set.}

\end{figure*}
 The best neural network trained on the canonical dataset had a loss
of 0.44 including regularization penalty and a mean square error of
0.22 and 0.41 for train and test set, respectively. The curves for
the training using the canonical dataset are very noisy (Figure 2A).
The best neural network trained on the enumerated dataset loss of
0.18 including regularization penalty and a mean square error of 0.09
and 0.30 for train and test set, respectively. The training curve
is significantly less noisy than for the canonical dataset (Figure
2B).

Both neural networks were used to predict the IC50 values from the
canonical and enumerated datasets, and the scatter plots are shown
in Figure 3. 

\begin{figure*}
\includegraphics[width=0.9\textwidth]{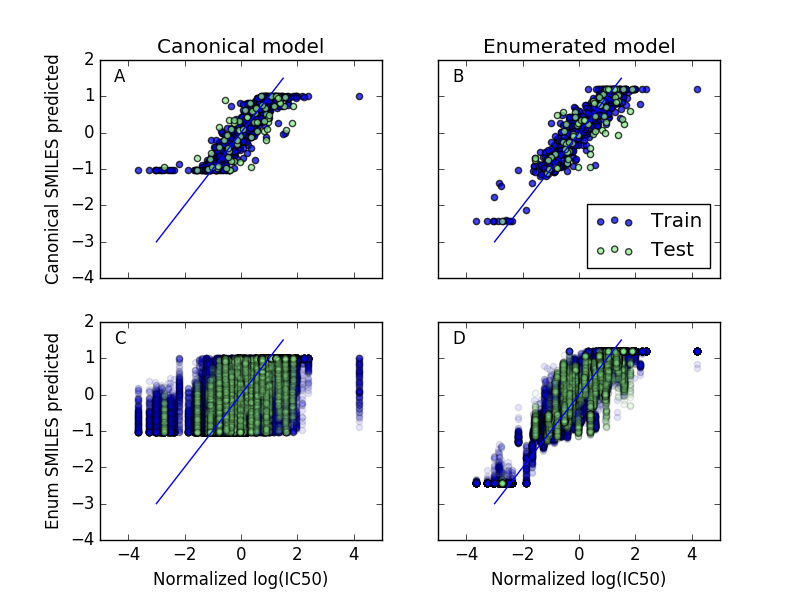}

\caption{Scatter plots of predicted vs. true values. Left column shows scatter
plots obtained with the model trained on canonical SMILES only. Right
column shows predictions with the model trained on enumerated data.
Top row is scatter plots with only canonical SMILES and bottom row
is predictions of the enumerated dataset. The blue line denotes the
perfect correlation (y = x).}
\end{figure*}
The correlation coefficients and root mean square deviation (RMS)
are tabulated in Table 3. The combination with the worst performance
was predicting the test set molecules is using enumerated SMILES neural
network model trained on the canonical dataset. Which has a correlation
coefficient of 0.26 and an RMS of 0.84. The bad correlation is clearly
visible from Figure 3 plot C. The best performance predicting the
test set, was seen with the combination of the enumerated model and
the enumerated SMILES. Here the correlation coefficient is 0.66 and
the RMS 0.55. The two other combinations, canonical model-canonical
SMILES and enumerated model, canonical SMILES are close in performance
(Table 3).\textsuperscript{2}

\begin{table}
\caption{Statistics of predicted values, values are for Train/Test set respectively }

\centering{}%
\begin{tabular}{cccccc}
 &  & \multicolumn{2}{>{\centering}p{1.3cm}}{{\small{}Canonical Model}} & \multicolumn{2}{>{\centering}p{1.3cm}}{{\small{}Enumerated Model}}\tabularnewline
\cline{3-6} 
\multicolumn{2}{c}{{\small{}Dataset}} & {\small{}R\textsuperscript{2}} & {\small{}RMS} & {\small{}R\textsuperscript{2}} & {\small{}RMS}\tabularnewline
\hline 
{\small{}Canonical} & {\small{}Train} & {\small{}0.78} & {\small{}0.46} & {\small{}0.85} & {\small{}0.39}\tabularnewline
 & {\small{}Test} & {\small{}0.56} & {\small{}0.62} & {\small{}0.63} & {\small{}0.56}\tabularnewline
\cline{2-6} 
{\small{}Enumerated} & {\small{}Train} & {\small{}0.25} & {\small{}0.88} & {\small{}0.87} & {\small{}0.37}\tabularnewline
 & {\small{}Test} & {\small{}0.26} & {\small{}0.84} & {\small{}0.66} & {\small{}0.55}\tabularnewline
\hline 
\end{tabular}
\end{table}

Figure 4 
\begin{figure}
\begin{centering}
\includegraphics[width=1\columnwidth]{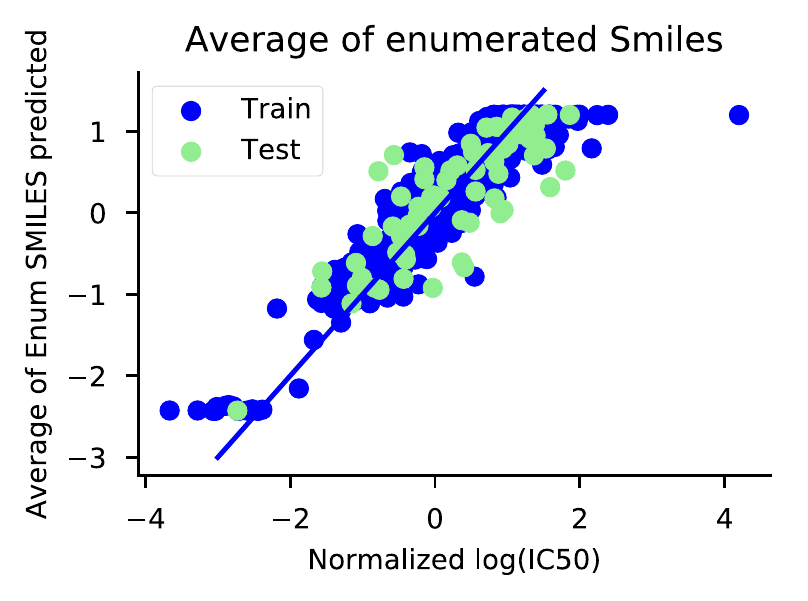}
\par\end{centering}

\caption{Average of predictions from the enumerated model for each molecule.
Train set R2 is 0.88 and RMS is 0.38. Test set R2 is 0.68 and RMS
is 0.52.}

\end{figure}
show a scatter plot of the average prediction for each molecule obtained
with the enumerated model. The calculated correlation coefficient
is 0.68 for the test set and the RMS is 0.52.

\section*{Discussion}

The results clearly suggest that SMILES enumeration as a data augmentation
technique for molecular data has benefits. The model trained on canonical
data is not able to predict many of the alternative SMILES of the
train and test set as is evident for Figure 3 plot C, where the bad
generalization to non canonical SMILES strings are evident. Instead
the best performance was observed by taking the average for each molecule
of the predictions of the enumerated SMILES using the enumerated model
(Figure 4), which shows that the SMILES enumeration can also be of
value in the sampling phase. The canonical model needed a lot more
epochs to train, but here it must be considered that the dataset contained
130 times fewer examples. Thus each epoch in the training was only
3 mini batches leading to 3 updates of the weights, whereas the enumerated
dataset had approximately 360 updates of the weights of the neural
network per epoch. The curves in Figure 2 thus represents 3000 and
18000 updates of the weights. The higher overhead of running more
epochs however led to approximately the same wall clock time in training.
The hyper parameters found during the optimization of the network
architecture and amount of regularization was not entirely as expected.
The expectation was that the canonical dataset would prefer a smaller
and simpler network with a larger regularization. Instead the canonical
dataset has a larger amount of LSTM cells (128) with no dropout, but
a much larger regularization of the final weights to the input neuron
(L1 and L2 maxed out at 0.2). The enumerated model had fewer LSTM
cells (64) and thus fewer connections, but nevertheless found dropout
on the input to the LSTM cells to be beneficial. To test if the differences
were due to the Bayesian optimization getting trapped in a local minimum,
the network architecture found for the enumerated dataset was test
trained with the dataset with the canonical SMILES only. The first
try with a learning rate of 0.005 failed (results not shown), but
lowering the learning rate to the one found for the canonical SMILES
(0.0001), gave a model with a correlation coefficient of 0.5 and RMS
of 0.68 on the train set. The predictive performance was even lower
with 0.45 and 0.69, for R2 and RMS respectively. The differences in
hyper parameters after optimization of using the two different datasets
thus seems justified. The study lacks the division into train, test
and validation set, where the hyper parameters are tuned on the test
set, but the final performance evaluated on the validation set. The
observed prediction performance of the LSTM-QSAR models are thus likely
overestimated to some degree. However, this study is focused on the
gains of using SMILES enumeration and not on producing the optimal
DHFR QSAR model. The performance on both the train and test set are
lower for the canonical model. If the differences in performance had
been due to to over-fitting, the smaller dataset would probably have
had an advantage. 

The use of SMILES as descriptors for QSAR is not new\cite{Worachartcheewan2014,DBLP:journals/corr/JastrzebskiLC16,Worachartcheewan2014}
and is as an example implemented in the CORAL software.\cite{Toropova2014}
The approach in the CORAL software is however very different from
the one in this study. CORAL software breaks down the SMILES into
single atoms, double atoms and triple atoms (Sk, SSk and SSSk) as
well as some extra manually coded extracted features such as BOND,
NOSP, HALO and PAIR.\cite{Toropova2014,Worachartcheewan2014} The
approach seems close to using a mixture of topological torsions\cite{Nilakantan1987}
with one, two and three atoms and atom-pair\cite{Carhart1985} fingerprints.
The LSTM-QSAR used in this approach directly uses the SMILES string
and supposedly let the model best extract the features from the SMILES
strings that best fit with the task at hand, and similar approaches
have been shown to outperform other common machine learning algorithms\cite{DBLP:journals/corr/JastrzebskiLC16},
although the details of optimization of the competing algorithms were
not completely clear.

SMILES were also used recently in an application of a neural network
based auto-encoder.\cite{Gomez-Bombarelli2016} Here the SMILES are
used as input to a neural network with the task of recreating the
input sequence. The information is passed through a “bottle-neck”
layer in between the encoder and the decoder, which limits the direct
transfer of information. The bottle neck layer thus ends up as a more
continuous floating point vector representation of the molecule, which
can be used to explore the chemical space near an input molecule,
interpolate between molecules and link the vector representation to
physico-chemical properties. The amount of unlabeled molecules for
the study already surpassed the needed amount, but could in principle
be expanded even more with the SMILES enumeration technique described
here. SMILES enumeration could possible allow the autoencoder to be
trained with smaller and more focused datasets of biological interest.
Additionally, tt would be interesting to see if different SMILES of
the same molecule ends up with the same vector representation or in
entirely different areas in the continuous molecular representations.

LSTM networks have also been used in QSAR applications demonstrating
learning transfer from large datasets to smaller.\cite{Altae-Tran2016}
Here the input was however not SMILES strings but rather molecular
graph convolution layers\cite{Duvenaud2015} working directly on the
molecular graph representation. The approach thus more directly reads
in the topology of the molecular model, rather than indirectly letting
the network infer the topology from the SMILES branching and ring
closures defined by the brackets and numbering in the SMILES strings.

\section{Conclusion}

This short investigation has shown promise in using SMILES enumeration
as a data augmentation technique for neural network QSAR models based
on SMILES data. SMILES enumeration enables the use of more limited
sizes of labeled data sets for use in modeling by more complex neural
network models. SMILES enumeration gives more robust QSAR models both
when predicting single SMILES, but even more when taking the average
prediction using enumerated SMILES for the same molecule. The SMILES
enumeration code as well as some of the scripts used for generating
the LSTM-QSAR models are available on GitHub: \href{https://github.com/Ebjerrum/SMILES-enumeration}{https://github.com/Ebjerrum/SMILES-enumeration}

\section{Conflicts of Interest}

E. J. Bjerrum is the owner of Wildcard Pharmaceutical Consulting.
The company is usually contracted by biotechnology/pharmaceutical
companies to provide third party services

\bibliographystyle{elsarticle-num}
\bibliography{NN_lit}

\end{document}